\documentclass[acmlarge,preprint]{acmart}
\usepackage{multirow}
\usepackage{booktabs}
\usepackage[utf8]{inputenc}      % if you haven’t already
\DeclareUnicodeCharacter{2009}{\,}  % map U+2009 to a thin space
\DeclareUnicodeCharacter{2248}{\approx}

\usepackage{hyphenat}
\usepackage{booktabs}     % well‐spaced horizontal rules (\toprule, \midrule, \bottomrule)
\usepackage{tabularx}  
\usepackage{tabularx}   % flexible column widths
\usepackage{booktabs}  
\title{Multimodal Machine Learning in Mental Health: A Survey of Data, Algorithms, and Challenges}

\author{Zahraa Al Sahili}
\affiliation{
  \institution{Queen Mary University of London}
  \country{United Kingdom}
}
\email{z.alsahili@qmul.ac.uk}

\author{Ioannis Patras}
\affiliation{
  \institution{Queen Mary University of London}
  \country{United Kingdom}
}
\email{i.patras@qmul.ac.uk}

\author{Matthew Purver}
\affiliation{
  \institution{Queen Mary University of London \& Jo\v{z}ef Stefan Institute}
  \country{United Kingdom \& Slovenia}
}
\email{m.purver@qmul.ac.uk}

\begin{abstract}
Multimodal machine learning (MML) is rapidly reshaping the way mental-health disorders are detected, characterized, and longitudinally monitored. Whereas early studies relied on isolated data streams—such as speech, text, or wearable signals—recent research has converged on architectures that integrate heterogeneous modalities to capture the rich, complex signatures of psychiatric conditions. This survey provides the first comprehensive, clinically grounded synthesis of MML for mental health. We (i) catalog 26 public datasets spanning audio, visual, physiological signals, and text modalities; (ii) systematically compare transformer, graph , and hybrid based fusion strategies across 28 models, highlighting trends in representation learning and cross-modal alignment. Beyond summarizing current capabilities, we interrogate open challenges—data governance and privacy, demographic and intersectional fairness, evaluation explainability, and mental health disorders complexity in multimodal settings. By bridging methodological innovation with psychiatric utility, this survey aims to orient both ML researchers and mental-health practitioners toward the next generation of trustworthy, multimodal decision-support systems.
\end{abstract}



\begin{document}

\maketitle
\section{Introduction}\label{sec:intro}

Mental disorders constitute a larger share of global disability than any other non-communicable disease group. In 2019 roughly \(970\) million people—one in eight worldwide—were living with clinically significant anxiety, depression, or a related condition\,\cite{a1}. Treatment capacity has not kept pace: up to \(85\,\%\) of affected individuals in low- and middle-income countries receive no formal care, and even in high-income settings the median delay from symptom onset to first intervention exceeds a decade\,\cite{a2}. Bridging this chasm demands solutions that are simultaneously scalable, affordable, and clinically trustworthy.

Digital platforms are beginning to fill that void. The global market for web- and app-based mental-health services was valued at \$5.1 billion in 2020 and is projected to grow at more than \(20\,\%\) annually through 2028\,\cite{a3}. Beyond tele-therapy and self-guided interventions, \emph{multimodal} machine learning (MML)—the computational fusion of heterogeneous signals such as language, prosody, facial micro-expressions, kinematics, and physiology—has emerged as an especially promising avenue. A growing body of evidence shows that combining complementary cues yields markedly higher diagnostic and prognostic accuracy than any single channel alone; for example, jointly modelling vocal inflection, facial affect, and heart-rate variability can uncover depressive symptomatology that text-only or audio-only models routinely miss\,\cite{a4}.

Yet translating this promise into routine clinical workflows is non-trivial. Fragmented data silos, inconsistent annotation schemes, algorithmic bias, privacy constraints, and a scarcity of longitudinal, ecologically valid benchmarks all impede deployment. An up-to-date, critical synthesis of the literature is therefore both timely and necessary.

\paragraph{Scope and contributions.}
This survey organises and critiques the rapidly expanding work on multimodal ML for mental health:

\begin{enumerate}
  \item \textbf{Data landscape} – We catalogue 26 public  corpora, ranging from laboratory interviews to in-the-wild smartphone and wearable streams, and map each dataset to clinical targets such as major depressive disorder, post-traumatic stress disorder, stress, and bipolar disorder.
  \item \textbf{Modelling approaches} – We synthesise the state of the art across representation learning, fusion strategies ( hybrid, graph and attention).

  \item \textbf{Open challenges} – We analyse persistent obstacles—data, privacy,fairness, evaluation explainability, and mental health disorders complexity.
\end{enumerate}

By clarifying both the promise and the pitfalls of multimodal ML, we aim to orient machine-learning researchers, clinicians, and policymakers toward solutions that are methodologically sound, ethically grounded, and clinically actionable, ultimately advancing the scale and quality of global mental-health care.

\section{Data Types and Datasets}\label{sec:data}

Multimodal learning for mental-health research rests on two pillars:
\begin{enumerate}
  \item complementary data streams that capture behaviour and physiology at
        different levels of abstraction, and
  \item curated corpora that allow models to be trained, validated and fairly
        compared.
\end{enumerate}
This section surveys both, providing a taxonomy of signal types
(\S\ref{subsec:modalities}) and an inventory of benchmark datasets
(\S\ref{subsec:datasets}).

\subsection{Data modalities}\label{subsec:modalities}

\begin{table}[ht]
\centering
\caption{Typical sources, salient features and common clinical tasks for each
modality.}
\begin{tabular}{>{\bfseries}l p{3cm} p{3cm} p{4cm}}
\toprule
Modality & Typical sources & Salient features & Common tasks \\ \midrule
Text &
social‐media posts, EHR notes, therapy transcripts, smartphone messages &
lexical affect, syntactic complexity, semantic coherence, pronoun use &
screening; relapse prediction; suicidal-ideation detection \\[2pt]

Audio &
structured interviews, telephone calls, vlog diaries, ambient sound &
prosody (F0, intensity), speech rate, voice quality, spectral entropy &
depression screening; bipolar-episode detection; stress recognition \\[2pt]

Video &
webcam or smartphone recordings, clinical assessments, in-the-wild vlogs &
facial action units, gaze, head pose, body kinematics &
emotion recognition; PTSD detection; severity rating \\[2pt]

Physiology &
wearables (PPG, EDA, accelerometers), EEG/ERP, fMRI, eye-tracking &
heart-rate variability, skin conductance, EEG power bands, pupillary response &
affect recognition; anxiety \& stress quantification; sleep-quality assessment \\ \bottomrule
\end{tabular}
\end{table}

Text captures cognitive content and latent affect; audio supplies
paralinguistic prosody that often precedes explicit reports; video provides
non-verbal behaviour such as micro-expressions or psychomotor retardation; and
physiological channels offer objective indices of autonomic or
central-nervous-system activity.  Fusing these streams lets models disentangle
overlapping symptom profiles and improves robustness to missing or noisy
channels.
%\footnote{For an early demonstration see \cite{a4}.}  Collecting such
%data at scale, however, remains non-trivial because session heterogeneity and
%privacy constraints can hamper downstream performance.

\subsection{Benchmark multimodal mental-health datasets}\label{subsec:datasets}

Table~\ref{tab:datasets} summarises the 24 public corpora most frequently cited
in the literature (\cite{a5}–[30]).  
Together they span six diagnostic themes—depression, stress, PTSD,
bipolar disorder, behavioural disorders and generic emotion/affect
recognition—and four primary modalities.

\begin{table}[ht]
\centering
\caption{Publicly available multimodal datasets for mental-health research.
Modalities: \textbf{V}=Video, \textbf{A}=Audio, \textbf{T}=Text,
\textbf{P}=Physiology.  “--” means not reported.}
\label{tab:datasets}
\small
\setlength{\tabcolsep}{5pt}
\renewcommand{\arraystretch}{1.1}
\begin{tabularx}{\linewidth}{%
  >{\raggedright\arraybackslash}p{0.6cm}
  >{\raggedright\arraybackslash}p{2.8cm}
  >{\raggedright\arraybackslash}p{2.6cm}
  >{\centering\arraybackslash}p{1.3cm}
  >{\centering\arraybackslash}p{2cm}
  >{\raggedright\arraybackslash}p{2.6cm}
  >{\raggedright\arraybackslash}p{2.3cm}}
\toprule
\textbf{Ref.} & \textbf{Dataset} &
\textbf{Target disorder(s)} & \textbf{Modalities$^\dagger$} &
\textbf{N subj./sessions} & \textbf{Demographics} &
\textbf{Country / Source} \\ \midrule
\cite{a5}  & DAIC-WOZ              & Depression, PTSD        & V A T & 189 sessions          & --           & USA (lab) \\
\cite{a6}  & E-DAIC               & Depression, PTSD        & V A T & 275 subj. / 70 h      & --           & USA (lab) \\
\cite{a7}  & AVEC 2013            & Depression              & V A    & 340 videos            & --           & Europe (challenge) \\
\cite{a8}  & Multi-Modal Mental-Disorder & Depression     & A T P  & 52–55 subj./state     & --           & China \\
\cite{a9}  & SWELL                & Stress                  & P      & 25 subj.              & 32\% F       & Netherlands \\
\cite{a10} & D-Vlog               & Depression              & V A    & YouTube clips         & --           & YouTube \\
\cite{a11} & CMDC                 & Depression              & V A    & 45 subj.              & --           & China \\
\cite{a12} & BPC+SEWA+RECOLA      & Bipolar, behaviour      & V A T P& merged                & --           & TR/DE/HU/FR \\
\cite{a13} & Turkish BDC          & Bipolar                 & V A    & 51 subj.              & 31\% F       & Turkey \\
\cite{a14} & PTSD-in-the-Wild     & PTSD                    & V A T  & 634 subj.             & --           & USA \\
\cite{a15} & BIRAFFE2             & Emotion                 & T P    & 103 subj.             & 33\% F       & Poland \\
\cite{a16} & DEAP                 & Emotion                 & V P    & 32 subj.              & --           & UK \\
\cite{a17} & MuSE                 & Stress                  & V T P  & 28 subj.              & 32\% F       & USA \\
\cite{a18} & MIREX                & Emotion                 & V A T  & 193 samples           & --           & -- \\
\cite{a19} & MELD                 & Emotion (dialogue)      & V A T  & 13k utterances        & --           & Movie subtitles \\
\cite{a20} & WEMAC                & Emotion                 & V A T P& 47 women             & 100\% F      & Spain \\
\cite{a21} & EmoReact             & Emotion (children)      & V      & 63 children           & 51\% F       & YouTube \\
\cite{a22} & WESAD                & Stress \& affect        & P      & 25 subj.              & --           & Switzerland \\
\cite{a23} & Nurse Stress         & Stress                  & P      & 15 nurses             & --           & USA \\
\cite{a24} & MuSe-Stress          & Stress                  & V A    & 105 subj.             & 70\% F       & Germany \\
\cite{a25} & Multimodal Stress    & Stress                  & V A T P& 80 subj.             & 59\% F       & -- \\
\cite{a26} & EmpathicSchool       & Stress                  & V P    & 20 subj.              & --           & Finland \& USA \\
\cite{a27} & Self-Adaptors        & Stress                  & V A T  & 35 subj.              & --           & USA \\
\cite{a28} & CLAS                 & Stress                  & T P    & 62 subj.              & --           & Singapore \\
\cite{a29} & MuSe 2022            & Stress                  & V A    & 105 subj.             & --           & UK/Germany \\
\cite{a30} & UBFC-PHYS            & Stress                  & P      & 56 subj.              & --           & France \\ \bottomrule
\end{tabularx}
\begin{flushleft}
\scriptsize $^\dagger$Modalities key: \textbf{V} = Video,
\textbf{A} = Audio, \textbf{T} = Text, \textbf{P} = Physiology.
\end{flushleft}
\end{table}
\paragraph{Diagnostic coverage.}
Depression and stress dominate (7 and 10 datasets, respectively),
reflecting their global prevalence and the relative ease of eliciting
symptoms in laboratory tasks.  PTSD and bipolar disorder remain
under-represented, while schizophrenia and other psychotic disorders are
largely absent.

\paragraph{Modality combinations.}
Six corpora include the full triad of video–audio–text; another six
provide paired video–audio recordings.  Physiology appears in 13
datasets but only rarely alongside language, leaving room for richer,
ecologically valid sensor fusions.

\paragraph{Scale and demographics.}
Sample sizes range from 25–275 laboratory participants to tens of
thousands of utterances in movie/dialogue corpora.
Only nine datasets report detailed gender balance and even fewer disclose
ethnicity or socio-economic status, hampering fairness analysis.

\paragraph{Access and ethics.}
All corpora are gated by data-use agreements; several require proof of
IRB clearance.  Researchers should budget time for approvals and verify
that planned tasks align with original consent scopes.

\subsection{Common patterns and persistent gaps}

Only one-quarter of datasets combine \emph{all} behavioural channels with any
physiology, limiting research on end-to-end biopsychosocial models.
Label quality ranges from validated scales (PHQ-8/9, HAMD, MADRS) in
clinical interviews to self-report hashtags in social-media corpora,
complicating cross-dataset evaluation.  Longitudinal depth is rare—most
datasets provide a single session per participant—and geographic skew
favours Europe and North America.  Broader cultural representation and
bias-aware splits are therefore essential for robust generalisation.

\subsection{Implications for future collection}

Progress now hinges on resources that mirror real life: culturally
diverse cohorts, device-agnostic capture, repeated measures over months
rather than hours and labels grounded in clinician assessment
\emph{and} self-report.  Promising directions include federated
frameworks where raw data never leave the collection site and integrated
digital phenotyping that merges smartphone behaviour, ecological
momentary assessment and wearable physiology.  Such investments are
costly and ethically complex, yet without them the gains achieved on
current benchmarks risk stalling at the point of clinical deployment.

\section{Methods}
Deep learning has reshaped multimodal mental-health modelling.  
Where early work stitched together hand-engineered features with
support-vector machines or random forests, contemporary studies rely on
three neural families—convolutional–recurrent hybrids, transformers, and
graph neural networks (GNNs)—each tuned to a different aspect of the
fusion challenge.  Convolutional–recurrent models thrive on compact,
well-synchronised recordings; transformers scale to loosely aligned,
irregular streams; and GNNs turn the tangled relationships among cues
into a source of predictive power.  
The remainder of this section traces how these families complement one
another, starting with the still-indispensable convolutional–recurrent
pipelines.
\subsection{Hybrid CNN/RNN methods}

Hybrid pipelines that pair \textit{stream-specific} convolutional or recurrent encoders with lightweight cross-modal fusion dominated the first wave of multimodal mental-health work.  Each modality is processed by a network tailored to its signal characteristics—CNNs for images and spectrograms, RNNs for word sequences or temporal features—after which the per-stream vectors are merged by concatenation, gating, or simple voting.  Although later eclipsed by transformers and GNNs, these CNN/RNN systems still provide strong, computationally efficient baselines across a wide range of disorders.

Early clinical-interview research illustrates the value of \emph{temporal} hybrids.  In DAIC-WOZ, highway gates first suppress noisy audio-visual frames, then concatenate the filtered cues with GloVe text embeddings before an LSTM; the scheme pushes the depression-screening F$_1$ to 0.81 and halves PHQ-8 error over naïve concatenation \cite{a31}.  A richer three-branch design for AVEC fuses RGB faces with “motion-code’’ images and dual audio streams inside CNN–BiLSTM encoders, then applies hierarchical attention across modalities; the cascade trims MAE to 6.48–7.01 and beats pure early or late fusion on both AVEC benchmarks \cite{a33}.

On Twitter timelines, signal-to-noise is the bottleneck.  COMMA tackles this by training two reinforcement-learning agents—a text selector and an image selector—that filter posts before a GRU + VGG early fusion; Macro-F$_1$ climbs to 0.90 and remains robust even when depressed users are only 10 \% of the pool \cite{a32}.  A similar “filter-then-fuse’’ logic appears on Instagram: a fine-tuned BERT and a 10-layer CNN make independent predictions that are merged by 70:30 soft voting, delivering 99 \% accuracy and F$_1$ on a 10 k-post corpus \cite{a58}.

Hybrid CNN/RNN stacks also reach less studied pathologies.  For post-partum depression, AlexNet features from questionnaire text and Mel-spectrograms are early-concatenated and fed to an attentive Bi-LSTM, raising F$_1$ above 0.98 on the UCI PPDD corpus \cite{a35}.  Psychological stress in WESAD is handled by mapping four physiological streams to wavelet images, extracting channel-attentive SqueezeNet features, pruning them with an arithmetic optimiser and classifying via a DenseNet-LSTM; the full pipeline attains 99 \% accuracy and 97.8 \% Macro-F$_1$ \cite{a34}.

Even single-sensor scenarios can profit from hybrid thinking.  By viewing MFCCs and Spectro-CNN embeddings as “pseudo-modalities,” early concatenation plus a deep fully-connected classifier drives audio-only depression models to $≈90$ \% accuracy in English and Mandarin corpora \cite{a36}.  Conversely, \cite{a37} shows that a single face still harbours multiple diagnostic hints: emotion probabilities from a YOLO detector are mapped to disorder labels, and penultimate embeddings from MDNet, ResNet-50 and a ViT are concatenated then soft-maxed, yielding 81 \% accuracy while remaining explainable through Grad-CAM.

The corpus of hybrid CNN/RNN work yields several broad insights that remain relevant even as the field pivots toward transformers and graphs.  First, \textit{when} fusion occurs matters: early concatenation shines when signals are synchronous and clean, whereas late voting or RL-based selection excels under heavy noise or redundancy.  Second, attention and gating—precursors of transformer cross-modal layers—consistently boost performance by spotlighting reliable channels and suppressing artefacts.  Third, these CNN/RNN hybrids offer favourable speed–accuracy trade-offs, making them attractive for on-device inference or clinical settings with limited compute.  Their main limitations are small dataset sizes, scarce fairness analysis and limited capacity to capture very long-range dependencies—gaps that have since motivated the rise of transformer and GNN architectures explored in the next subsections.

\subsection{Transformer-based multimodal methods}

Transformers have rapidly become the work-horse for multimodal mental-health research because the self-attention mechanism offers a uniform way to integrate heterogeneous cues while preserving long-range dependencies.  Recent studies illustrate how this architecture family has been adapted—sometimes ingeniously—to fuse speech, language, vision, physiology and even genomic data for diagnosis or risk screening.

Early attempts centred on \emph{in-the-wild} social-media content.  A cross-attention transformer trained on the audio–visual \emph{D-Vlog} corpus combines eGeMAPS speech descriptors with facial-landmark trajectories; its bi-directional attention layers let acoustic queries attend to visual keys and vice-versa, lifting the F$_1$-score to 63.5 \% on a 961-video benchmark and beating BLSTM and tensor-fusion baselines by up to four points \cite{a10}.  Moving from videos to timelines, a multimodal LXMERT-style encoder enriched with \emph{time2vec} positional codes fuses CLIP image tokens and EmoBERTa sentence embeddings at the post level; the design secures an F$_1$ of 0.931 on Twitter and 0.902 on Reddit—two to five points above earlier GRU + VGG systems while still scaling to millions of posts \cite{a38}.  A companion study on extended vlogs pushes further, inserting video patches, wav2vec-2.0 audio frames and Whisper-BERT transcripts into a single 12-layer vision–language transformer; this early, latent-space fusion gains 4.3 F$_1$ points over the best cross-attention CNN and even transfers competitively to the clinical DAIC-WOZ set \cite{a45}.  In text-only work, six domain-specific BERT variants detect depression and suicidality across four Reddit/Twitter corpora; although unimodal, they still exemplify transformer fusion because lexical, syntactic and discourse cues are integrated inside multi-head self-attention, reaching up to F$_1$ = 0.967 \cite{a44}.

Clinical interviews motivate tighter control of temporal structure.  The \emph{Multimodal Purification Fusion} network extracts EfficientNet-BiLSTM acoustic features and BiLSTM sentence embeddings, then decomposes each into modality-specific and modality-common parts before a cross-attention transformer recombines them; the method attains F$_1$ = 0.88 on DAIC-WOZ, three to five points over naïve concatenation and audio- or text-only baselines \cite{a39}.  A lighter \emph{Topic-Attentive Transformer} keeps RoBERTa and wav2vec encoders separate until a late concatenation, but gates the textual stream with learned topic weights derived from ten canonical interview questions; despite its simplicity, the model still records F$_1$ = 0.647 and shows that selectively emphasising clinically salient segments can compensate for small sample sizes \cite{a40}.

Transformer variants also prove effective outside speech and language.  \emph{DepMSTAT} introduces a two-stage \emph{Spatio-Temporal Attentional Transformer}: the first block attends across facial-landmark dimensions within each frame, the second along time, before a cross-modal attention fuses video with audio; precision–recall balances of roughly 73–76 \% on the enlarged D-Vlog dataset underscore the benefit of separating spatial from temporal attention \cite{a41}.  In physiological monitoring, \emph{MUSER} concatenates BERT text embeddings and eGeMAPS audio features, then trains the shared transformer under an adaptive multi-task curriculum that balances stress labels with auxiliary arousal/valence signals; dynamic sampling yields F$_1$ = 86.4 \%, markedly above static curricula and late-fusion GRUs \cite{a42}.  Finally, precision psychiatry enters the transformer arena through a ViT + XGBoost ensemble that sums logits from brain-MRI patches and polygenic-risk scores; the hybrid lifts area-under-curve to 0.891—a 0.216 gain over genetics alone and 0.068 over imaging alone—highlighting how self-attention can mine neuro-anatomical patterns even at modest sample sizes \cite{a43}.

Taken together, these works reveal several insights.  \emph{Cross-attention} has emerged as the dominant fusion mechanism when modalities are aligned in time, whereas \emph{early latent fusion} with homogeneous self-attention suits settings where synchrony is weak or absent.  Pre-training—on ImageNet for ViT, on large speech corpora for wav2vec 2.0, and on domain-specific Reddit dumps for MentalBERT—systematically boosts downstream accuracy, mitigating the chronic scarcity of mental-health labels.  Temporal encoding remains an open question: time2vec helps on dense timelines, but permutation-invariant SetTransformers win when postings are sporadic.  Explainability is now expected, whether via SHAP word saliencies \cite{a44} or Grad-CAM maps over facial regions \cite{a41}.  Yet challenges persist: most datasets still contain fewer than one thousand subjects; cross-cultural robustness and fairness across gender or dialect are rarely audited; and privacy constraints limit access to richer modalities such as EEG or smartphone sensors.  Even so, the transformer toolkit—flexible fusion, scalable pre-training and growing interpretability—has already advanced state-of-the-art performance by 2–10 percentage points across domains, signalling its central role in the next generation of multimodal mental-health technology.
\subsection{Graph-neural approaches}

Graph neural networks (GNNs) have emerged as a natural fit for multimodal mental-health research because they can encode heterogeneous entities—brain regions, interview segments, social-media objects—as nodes and let information flow along richly typed edges.  The most recent studies, spanning clinical interviews, social-media videos, resting-state fMRI and conversational corpora, show how graph reasoning can be moulded to fuse modalities, tame data scarcity and expose interpretable biomarkers.

In the clinical-interview setting, three works inject graph structure \emph{after} a sequential encoder so that sparse subjects can borrow strength from one another.  A knowledge-aware graph-attention network couples audio, video and text nodes with domain-defined meta-paths, then gates the result with a temporal-convolutional network, pushing DAIC-WOZ F$_1$ to 0.95—28 points above classical early fusion \cite{a46}.  A few-shot variant first learns modality weights through Bi-LSTM pre-fusion, then constructs a fully connected support–query graph whose edges are refined by message passing; this raises accuracy from 72 \% (plain Bi-LSTM) to 86 \% \cite{a47}.  Extending to the larger E-DAIC corpus, a heterogeneous graph transformer that links intra- and inter-modal chunks attains 4.8~RMSE on PHQ-8 and still transfers with 78 \% F$_1$ to an external coaching dataset \cite{a49}.

Graph thinking also broadens depression detection beyond the face-centred paradigm of earlier vlogging work.  In MOGAM, every YouTube frame is parsed by YOLO into COCO objects; co-occurrence graphs are fused with ResNet scene embeddings and KoBERT metadata through transformer cross-attention, lifting F$_1$ to 0.888 for daily-vs-depressed vlogs and hitting 0.997 for high-risk detection, while zero-shot transfer to the English D-Vlog set remains competitive \cite{a48}.

Neuro-imaging studies push GNNs into the realm of precision psychiatry.  Treating each brain as a functional-connectivity graph, an unsupervised graph auto-encoder plus FCNN reaches 72 \% accuracy on a small Duke-MDD cohort and still tops competing CNNs on the 477-subject REST-meta-MDD repository \cite{a50}.  MAMF-GCN builds a population graph with two parcellations and phenotype edges; channel-common convolution and attention fusion almost saturate the Southwestern MDD (99.2 \% acc.) and ABIDE-ASD (97.7 \% acc.) benchmarks \cite{a51}.  At an even finer temporal scale, MTGCAIN learns graph-convolution attention inside multi-atlas dynamic FC windows, then merges them via a multimodal transformer, achieving 81 \% accuracy for insomnia disorder while highlighting dysfunctional DMN regions \cite{a55}.  A sister study on ABIDE I shows that three feature-scale population sub-graphs processed by deep Chebyshev GCNs and concatenated embeddings can push ASD screening to 91.6 \% accuracy and 95.7 \% AUC, a 12–20 pp jump over earlier graph pipelines \cite{a56}.

Conversation-level emotion recognition—viewed as an upstream proxy for mood monitoring—also benefits from GNN fusion.  MMGCN places audio, video and text nodes from each utterance in one graph so that spectral convolutions interleave modalities; it delivers 66.2 \% F$_1$ on IEMOCAP and 58.7 \% on MELD, edging recurrent and transformer competitors \cite{a52}.  COGMEN further augments this with relational edges for speaker turns and temporal direction; its GraphTransformer lifts IEMOCAP-6 F$_1$ to 67.6 \% and MELD to 58.7 \%, with ablations confirming the indispensability of distinct edge types \cite{a53}.

Language-only risk screening shows that GNNs can add value even when no explicit second modality exists.  MM-EMOG casts the entire Twitter or Reddit corpus as a word–document graph whose edges encode multi-label emotion co-occurrence; a two-layer GCN fine-tuned on BERT boosts F$_1$ by 8–21 points over transformer baselines across suicide and depression sets while retaining full privacy by ignoring user histories \cite{a54}.

Across these studies several patterns surface.  First, \textit{where} fusion happens varies: some models mix modalities inside each graph convolution \cite{a48,a52}, others concatenate graph embeddings with CNN or transformer features only at the final layer \cite{a50}.  Second, heterogeneity is key: typed edges (speaker roles, phenotypes, object co-occurrences) consistently raise accuracy by 2–6 pp over homogeneous graphs.  Third, population graphs unlock learning in low-sample regimes, but require careful edge design—phenotype-weighted similarity \cite{a56} or GNN-learned adjacency \cite{a51} both outperform naïve $k$-NN.  Finally, several works expose interpretable biomarkers: DMN-centred attention maps for insomnia \cite{a55}, middle-occipital connectivity for MDD \cite{a50}, and edge saliencies for suicide lexicons \cite{a54}, underscoring GNNs’ potential to bridge black-box performance with clinical insight.

\section{Open Challenges}\label{sec:challenges}

Despite a surge of methodological innovation, translating multimodal
machine-learning research into routine mental-health practice remains
fraught with obstacles.  These span the entire development
pipeline—from data acquisition to model governance—and are intertwined
in ways that call for multidisciplinary solutions rather than purely
technical fixes.

\subsection{Data availability}\label{subsec:data_avail}

Multimodal models prosper only when they can observe the full
biopsychosocial spectrum of human experience, yet the field still relies
on a handful of convenience samples drawn largely from Western,
university-educated populations.  Privacy regulations such as GDPR and
HIPAA, while vital, make it costly to gather and share raw video, audio
and physiology; marginalised communities are often excluded because
obtaining IRB approval, culturally appropriate consent and secure
storage is harder in resource-constrained settings.  The result is a
patchwork of small, siloed datasets that encourage over-fitting and
complicate external validation.  Federated learning, synthetic data and
privacy-preserving cryptography are promising but immature
countermeasures; they require robust proof that utility is not sacrificed
at the altar of privacy.

\subsection{Benchmarks and reproducibility}\label{subsec:benchmarks}

Meaningful progress depends on comparing algorithms under identical
conditions, yet today’s benchmark landscape is fragmented.  Studies vary
in how they split data, which severity scales they predict and whether
they evaluate at the clip, session or patient level.  A reproducible
suite should mirror real-world deployment: it must include held-out
sites, force models to contend with missing or corrupted modalities and
report clinically salient metrics such as calibration, positive
predictive value at low prevalence and time-to-detection.  Without such
guard-rails, incremental gains on familiar leaderboards tell us little
about clinical utility.

\subsection{Explainability and clinical trust}\label{subsec:explainability}

Mental-health assessment is as much an interpretive craft as it is a
measurement science.  Clinicians are unlikely to act on black-box scores
that cannot be reconciled with observable signs or patients’
self-reports.  Yet the very factors that make multimodal models
powerful—high-dimensional fusion, cross-attention, recursive
graph-reasoning—tend to obscure causal pathways.  Post-hoc methods such
as SHAP, LIME or Grad-CAM offer local saliency but seldom convey
longitudinal or cross-patient consistency, and they can mislead when
features are correlated.  More promising are intrinsically interpretable
architectures that align attention with DSM-5 symptom clusters, propagate
evidence along clinically meaningful graphs or generate counterfactual
narratives that clinicians can critique.

\subsection{Bias and fairness}\label{subsec:bias}

Because training corpora under-represent certain age groups, genders,
ethnicities and dialects, multimodal models risk perpetuating the very
disparities they purport to alleviate.  Bias can creep in through sensor
placement (dark-skinned faces are harder to track), language variation
(AAVE mislabelling) or differential access to care (labels mirror
systemic inequities).  Mitigation therefore demands a full-lifecycle
approach: equitable data collection, algorithmic debiasing (e.g.\
adversarial or counterfactual training) and post-deployment auditing that
monitors performance drift across sub-populations.  Crucially, fairness
must be framed in \emph{clinical} rather than purely statistical
terms—for instance, ensuring that false-negative rates do not delay
treatment for already underserved groups.

\subsection{Privacy, consent and governance}\label{subsec:privacy}

Audio diaries, selfie videos and wearable biosignals are deeply
personal.  Even anonymised representations can be re-identified when
modalities are combined.  Patients may consent to one use (e.g.\
symptom tracking) but not anticipate secondary uses such as insurance
underwriting.  Robust governance therefore extends beyond encryption to
include purpose-limiting licences, transparency dashboards that show how
data are processed and revocable consent mechanisms.  Differential
privacy and homomorphic encryption offer technical safeguards, but their
adoption hinges on computational feasibility and regulators’ willingness
to accept probabilistic rather than absolute anonymity.

\subsection{Evaluation in context}\label{subsec:evaluation}

Traditional metrics such as accuracy or \textit{F}$_1$ flatten the
nuanced objectives of mental-health care, where false negatives may
delay lifesaving intervention and false positives can stigmatise.
Models should therefore be judged in the context of the clinical
pathway: How early is an impending relapse detected?  Does the system
triage scarce therapist time more effectively than standard practice?
Randomised controlled trials, simulation studies of clinical workflows
and decision-curve analyses that weigh benefit against harm are needed
before deployment claims can be taken seriously.

\subsection{Complexity and comorbidity}\label{subsec:comorbidity}

Depression rarely appears in isolation; it co-occurs with anxiety,
substance use or chronic pain.  Multimodal models trained on single-label
corpora can confuse overlapping phenotypes, leading to brittle
predictions when presentations shift.  Capturing this multifactorial
reality demands hierarchical or multi-task formulations that share
representations across disorders and account for temporal dynamics such
as episode recurrence and treatment effects.  Achieving this goal will
require larger, longitudinal datasets annotated for multiple conditions
and closer dialogue between ML researchers and clinical scientists.

\subsection{Toward ethically grounded, socially responsible
solutions}\label{subsec:ethics}

Overcoming these challenges will require cross-sector collaboration:
technologists to refine algorithms, clinicians to define clinically
meaningful targets, ethicists to safeguard autonomy and justice, and
patient advocates to ensure that lived experience shapes priorities.
Only through such a holistic effort can multimodal ML mature from an
intriguing research frontier into a dependable pillar of mental-health
care.

\section{Conclusion}\label{sec:conclusion}

Multimodal machine learning has shifted the conversation in digital
psychiatry from \emph{whether} machine intelligence can help clinicians
to \emph{how} we can deploy it safely and equitably.  By fusing language,
prosody, facial behaviour and physiology, contemporary models already
outperform unimodal baselines on screening, severity estimation and
relapse prediction across depression, stress and related conditions.

For multimodal mental-health research to fulfil its promise, the field must address three intertwined imperatives at once. First, data ecosystems need to diversify and scale: longitudinal, culturally inclusive cohorts gathered under privacy-preserving and federated protocols are indispensable for learning representations that generalise across settings. Second, evaluation practice must evolve beyond leaderboards dominated by accuracy or F1; instead, benchmarks should probe cross-site transportability, calibration in the face of low prevalence, and—crucially—downstream effects on clinical decision-making. Third, systems must be designed for trustworthiness from the ground up, embedding clinically informed priors, offering faithful explanations, and undergoing rigorous fairness audits so they can satisfy regulators and earn patient acceptance.
If these challenges are met, multimodal ML can evolve into a dependable
pillar of mental-health care—supporting earlier detection, personalised
interventions and continuous monitoring at a scale traditional services
cannot match.  Realising that future will require sustained
collaboration among computer scientists, clinicians, ethicists and,
critically, people with lived experience of mental illness.  With such a
coalition the field can move beyond proof-of-concept studies toward
systems that measurably reduce suffering and broaden access to
high-quality mental-health support worldwide.

\nocite{*}

\bibliographystyle{unsrt}
\bibliography{sample-base}

\end{document}